\documentclass[letterpaper, 10 pt, conference]{ieeeconf}  

\IEEEoverridecommandlockouts                              

\usepackage{graphics} 
\usepackage{epsfig} 
\usepackage{times} 
\usepackage{amsmath} 
\usepackage{amssymb}  
\usepackage{booktabs}
\usepackage{algorithm}
\usepackage{algpseudocode}
\usepackage{multirow}
\usepackage{tabularx}
\usepackage{float}
\usepackage{authblk}
\usepackage{cite}
\makeatletter
\let\NAT@parse\undefined
\makeatother
\usepackage[colorlinks=true, linkcolor=blue, citecolor=blue, hypertexnames=true]{hyperref}
\usepackage{xcolor,colortbl}
\definecolor{lgreen}{RGB}{236, 255, 201}
\definecolor{nvgreen}{RGB}{118, 185, 0}
\title{\LARGE \bf
Enhancing Autonomous Driving Safety with \\
Collision Scenario Integration
}

\def\authorBlock{
    Zi Wang\textsuperscript{1, 2}\textsuperscript{*\dag}\thanks{*\,Work done during an internship at NVIDIA.}\thanks{\dag\,Corresponding author: \url{ziwang@nvidia.com}.},
    Shiyi Lan\textsuperscript{2},
    Xinglong Sun\textsuperscript{2},
    Nadine Chang\textsuperscript{2},
    Zhenxin Li\textsuperscript{3},
    Zhiding Yu\textsuperscript{2}, 
    Jose M. Alvarez\textsuperscript{2}  \\
  {$^1$Carnegie Mellon University} \quad {$^2$NVIDIA} \quad {$^3$Fudan University} 
}

\author{\authorBlock}

\begin{document}

\newcommand{\ja}[1]{{\color{red}Jose:  #1}}
\newcommand{\as}[1]{{\color{green}Alex:  #1}}
\newcommand{\syl}[1]{{\color{cyan}Shiyi: #1}}
\newcommand{\nadine}[1]{{\color{magenta}Nadine: #1}}
\newcommand{\ziwang}[1]{{\color{blue}Zi:  #1}}

\newcommand{\para}[1]{\medskip\noindent\textbf{#1.}}
\maketitle
\thispagestyle{empty}
\pagestyle{empty}

\begin{abstract}

Autonomous vehicle safety is crucial for the successful deployment of self-driving cars. However, most existing planning methods rely heavily on imitation learning, which limits their ability to leverage collision data effectively. Moreover, collecting collision or near-collision data is inherently challenging, as it involves risks and raises ethical and practical concerns. In this paper, we propose SafeFusion, a training framework to learn from collision data. Instead of over-relying on imitation learning, Safefusion integrates safety-oriented metrics during training to enable collision avoidance learning. In addition, to address the scarcity of collision data, we propose CollisionGen, a scalable data generation pipeline to generate diverse, high-quality scenarios using natural language prompts, generative models, and rule-based filtering. Experimental results show that our approach improves planning performance in collision-prone scenarios by 56\% over previous state-of-the-art planners while maintaining effectiveness in regular driving situations. Our work provides a scalable and effective solution for advancing the safety of autonomous driving systems.

\end{abstract}

\section{INTRODUCTION}

Improving the safety of autonomous vehicles is a critical priority, particularly in unpredictable and hazardous situations. However, real-world deployment presents two major challenges. First, acquiring collision data is inherently difficult since collecting those data and recreating dangerous situations poses significant risks, which might lead to ethical and practical constraints. Second, effective learning of collision data~\cite{bansal2018chauffeurnet, zhou2021exploring, zeng2019end, cui2021lookout, cheng2024rethinking, lu2023imitation, liu2022improved, huang2022efficient, li2024think2drive, casas2021mp3, gu2023vip3d, hu2023planning, jiang2023vad, chen2024vadv2, sadat2020perceive, ye2023fusionad, li2024ego, shao2023safety, li2024hydra} remains a challenge. Collision scenario data often lack avoidance trajectories, and integrating regular and collision data for training can degrade performance due to data imbalance and domain discrepancies. These obstacles significantly hinder progress in autonomous driving safety, which requires innovative approaches for collision data generation and planning training using collision data. 

Various scene generation methods~\cite{philion2024trajeglish, wu2024smart, zhao2024kigras, feng2023trafficgen} have been proposed to synthesize realistic traffic. However, these approaches struggle to generate diverse collision scenarios due to the limited number of such cases in their training datasets~\cite{ettinger2021large} and the inherent lack of controllability in their architectural designs. To address this limitation, we introduce \textit{CollisionGen}, a scalable collision data generation pipeline consisting of three key components: a tailored prompt system, a generative model, and a rule-based filtering process. We utilize LCTGen~\cite{tan2023language} as the generative model and develop a customized prompt system that provides precise collision scenario descriptions, enhancing controllability and realism. Additionally, we introduce a rule-based filtering and simulation process to ensure all generated data align with the intended prompt inputs, maintaining consistency and quality. This pipeline is a scalable and efficient approach for generating realistic collision scenarios that adhere to physical laws and meet user-defined requirements.

\begin{figure}[t]
    \centering
    \includegraphics[width=0.43\textwidth]{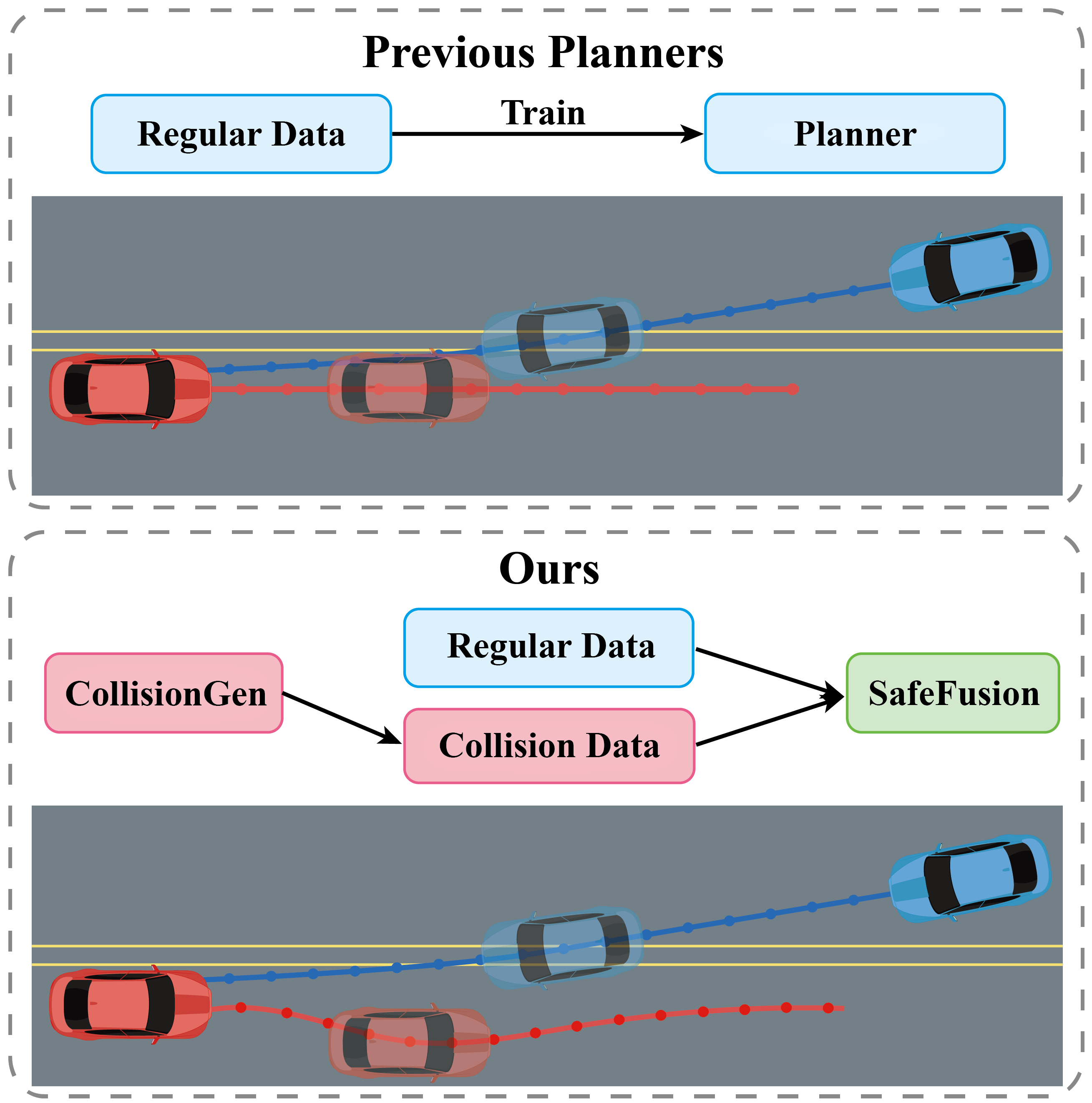} 
    \caption{The {\color{red}red car} represents the ego vehicle controlled by the planners, while the {\color{blue}blue car} represents the other vehicle. Previous planners struggle with dangerous scenarios due to the absence of collision data in real-world regular datasets. With \textit{CollisionGen} and \textit{SafeFusion}, planners achieve effective collision avoidance. }
    \label{fig:teaser}
    \vspace{-1em}
\end{figure}

Even with access to collision data, regardless of whether it is collected from real-world scenarios or generated artificially, integrating it into planner training presents significant challenges. First, collision scenarios data lack the trajectories that successfully avoid collisions, making it difficult for planners to learn effective avoidance strategies. This limitation restricts the use of imitation learning, a common approach in training neural planners~\cite{bansal2018chauffeurnet, zhou2021exploring, zeng2019end, cui2021lookout, cheng2024rethinking, lu2023imitation, liu2022improved, huang2022efficient, li2024think2drive, casas2021mp3, gu2023vip3d, hu2023planning, jiang2023vad, chen2024vadv2, sadat2020perceive, ye2023fusionad, li2024ego, shao2023safety, li2024hydra}. Second, directly integrating regular and collision data during training can degrade performance due to imbalanced data distributions and domain discrepancies~\cite{saerens2002adjusting, ben2010theory}. Unlike prior works~\cite{feng2023trafficgen}, which focus on general synthetic data, training with collision scenarios introduces unique challenges that remain unaddressed.

To tackle the difficulties in training with both regular and collision data, we propose \textit{SafeFusion}, a training framework that leverages a planning trajectory vocabulary and a multi-target knowledge distillation, inspired by~\cite{li2024hydra}. First, we construct a trajectory vocabulary~\cite{chen2024vadv2, li2024hydra} by clustering ego vehicle trajectories from real-world data, providing the planner with structured motion patterns. Then, we apply multi-target knowledge distillation, using offline simulation scores—evaluating safety, efficiency, and comfort—as supervisory signals to guide trajectory selection. These components eliminate neural planner's reliance on imitation learning. To balance training, we sample batches proportionally from both datasets and apply adaptive loss weights~\cite{saerens2002adjusting, ben2010theory} to mitigate data imbalance and domain gaps. Our framework can effectively handle collision data, whether collected from real-world environments or generated synthetically. In this work, due to the lack of access to real-world collision datasets, we utilize the synthetic data \textit{CollisionGen} generated for both training and testing, as shown in Fig.~\ref{fig:teaser}.

To summarize, the main contributions of our work are: 
\begin{itemize} 
\item We present a scalable collision data generation pipeline \textit{CollisionGen} for generating realistic collision scenarios using natural language prompts. Using it, we created \textit{Collision2k}, a dataset of approximately 2,000 diverse ego-vehicle crashes from various directions under different road conditions. This dataset facilitates training and testing of planning algorithms.  
\item We propose an effective training framework \textit{SafeFusion} that integrates collision data into the training of neural-based planners, tackling reliance on imitation learning, data imbalance and domain gap challenges.
\item We experimentally demonstrate that our framework significantly enhances planning performance in hazardous scenarios while maintaining effectiveness on real-world regular datasets, achieving up to \textbf{63.5\%} improvement in safety metrics and \textbf{56\%} in overall performance.
\end{itemize}

\vspace{-3pt}
\section{RELATED WORK}
\subsection{Traffic Modeling and Scene Generation}

Traffic modeling aims to create controlled environments that mimic real-world conditions and generate critical scenarios at scale. Traditionally, traffic scenario generation has relied on expert-defined rules~\cite{lopez2018microscopic, papaleondiou2009trafficmodeler, maroto2006real, gunawan2012two, erdmann2015sumo}, widely used in virtual driving datasets~\cite{wrenninge2018synscapes, ros2016synthia} and simulators~\cite{dosovitskiy2017carla, prakash2019structured}. However, these rule-based methods struggle to capture the complexity of real-world traffic scenarios. Learning-based approaches like SMART~\cite{wu2024smart}, KiGRAS~\cite{zhao2024kigras}, and BehaviorGPT~\cite{zhou2024behaviorgpt}, as seen in Waymo's Sim Agent Challenge~\cite{Montali2023neurips_wosac}, simulate future agent motions based on historical data. SceneGen~\cite{tan2021scenegen} uses auto-regressive models for scenario generation, while TrafficGen~\cite{feng2023trafficgen} applies modules for agent initialization and motion prediction. However, these methods rely on random scenario generation and lack customization, making it difficult to reproduce specific corner cases.
LCTGen~\cite{tan2023language} improves scenario customization by converting natural language prompts into structured scenarios using ChatGPT~\cite{openai2023gpt} and a generative transformer, but it remains inefficient in designing critical corner cases, reducing its effectiveness for planning integration. In contrast, our \textit{CollisionGen} generates diverse, high-risk collision scenarios through a tailored prompt system and rule-based filtering. The resulting dataset, \textit{Collision2k}, is designed for training and evaluating planners in safety-critical situations.

\vspace{-2pt}
\subsection{Planning in Autonomous Driving and Its Safety}
Planning algorithms are crucial for ensuring efficient vehicle operation, with extensive research exploring both post-perception and end-to-end motion planning approaches. Traditional post-perception planners are rule-based~\cite{treiber2000congested, thrun2006stanley, leonard2008perception, urmson2008autonomous, fan2018baidu, dauner2023parting}, offering structured, interpretable, and reliable decision-making framework. In contrast, learning-based post-perception planners leverage data-driven techniques such as imitation learning~\cite{bansal2018chauffeurnet, zhou2021exploring, zeng2019end, cui2021lookout, cheng2024rethinking} and reinforcement learning~\cite{lu2023imitation, liu2022improved, huang2022efficient, li2024think2drive}, making them more adaptable to complex driving scenarios and scalable with data. Another emerging trend in autonomous driving is end-to-end planning~\cite{casas2021mp3, gu2023vip3d, hu2023planning, jiang2023vad, chen2024vadv2, sadat2020perceive, ye2023fusionad, li2024ego, shao2023safety, li2024hydra}, where raw sensor data is directly processed to generate planning outputs. Among these, Hydra-MDP~\cite{li2024hydra} represents the state-of-the-art, integrating imitation learning and knowledge distillation to guide the planner in selecting trajectories from a predefined trajectory vocabulary. In our work, we adopt and modify this architecture by taking post-perception bounding boxes as input and removing the imitation learning component. This modification addresses the challenge of lacking collision-avoidance trajectories in the collision dataset, ensuring the planner learns effective collision mitigation strategies. Moreover, most existing planners do not explicitly prioritize safety. Current methods for enhancing safety in autonomous driving primarily rely on simulations and rule-based safety checks~\cite{vitelli2022safetynet, pini2023safe}. In contrast, our approach pioneers a data-driven method to improve safety by incorporating collision scenarios into the training process, enabling planners to better handle high-risk situations.

\vspace{-3pt}
\section{Method } 

\begin{figure*}[htbp]
    \centering
    \includegraphics[width=\textwidth]{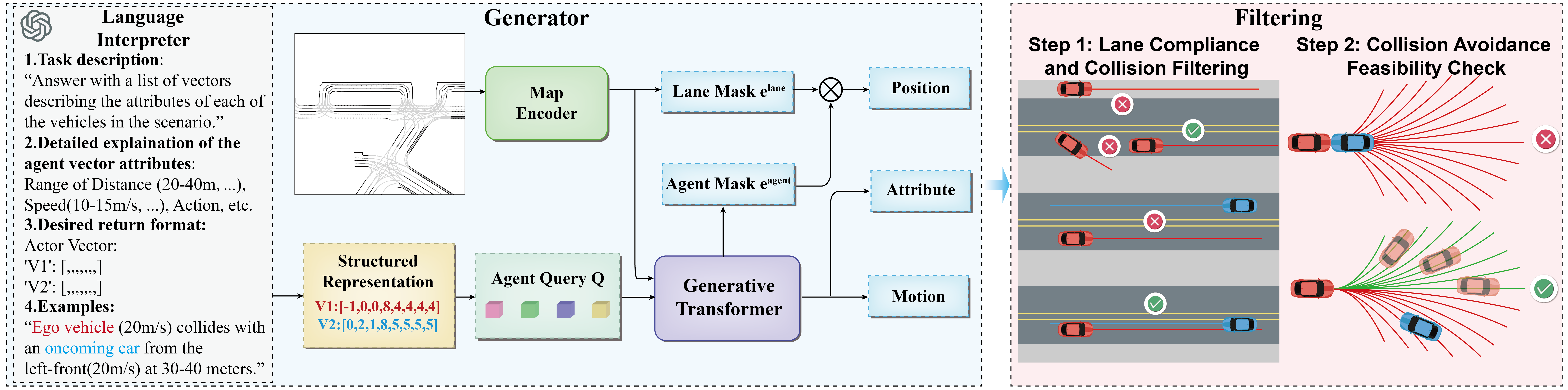} 
    \caption{The pipeline begins by taking text descriptions of collision scenarios as input. A generator with a language interpreter and a generative transformer is then applied, followed by the use of predefined rules and a PDM simulator~\cite{dauner2023parting} to filter out qualified collision scenarios. These filtered scenarios are subsequently used for the training and evaluation processes of planners. }
    \label{fig:generation_pipeline}
    \vspace{-2em}
\end{figure*}


In this section, we first introduce the scalable \textit{CollisionGen} pipeline for generating realistic crash scenarios in Sec.\ref{sec:data_generation}, covering the generator architecture and post-processing steps. We then describe \textit{SafeFusion}, our training framework that effectively trains planners using both real-world regular dataset and collision dataset, leveraging a planning trajectory vocabulary, multi-target knowledge distillation, and a data-balancing strategy, as discussed in Sec.\ref{sec:planning}.

\vspace{-2pt}
\subsection{CollisionGen}
\label{sec:data_generation}

Fig.~\ref{fig:generation_pipeline} illustrates the scalable data generation pipeline. \textit{CollisionGen} generates realistic collision scenarios that comply with user-defined requirements and physical laws, based on a textual description $C$ and a map dataset $M$. Each map $m\in M$ represents a localized road network region composed of $S$ lane segments, $m = (v_1, v_2, ..., v_S)$. Each lane segment $v_j$ is defined as a directed road section with endpoints $(x_{1j}, y_{1j})$ and $(x_{2j}, y_{2j})$, representing the start and end points of the segment. The pipeline is composed of two key components: (1) a \textbf{Generator} that synthesizes a scenario from the description $C$ and the map $m$, and (2) a \textbf{filtering} procedure that effectively selects diverse collision scenarios.

\subsubsection{\textbf{Generator}}
We adopt the LCTGen~\cite{tan2023language} framework but address its reliance on redundant and ambiguous Crash Report~\cite{crashreport} text by designing specialized prompts for collision scenarios. These prompts depict the ego vehicle driving normally while other vehicles collide from various directions. A simple example is shown in Fig.~\ref{fig:generation_pipeline}.

The Generator begins with a Language Interpreter, which processes the collision text description $C$ and produces a structured representation $z$ via GPT-4o~\cite{openai2023gpt}.
Following LCTGen~\cite{tan2023language}, the structured representation $z=[V_1, ..., V_N]$ contains 8-dimensional vectors for each agent, detailing an agent's quadrant position index ($1-4$), distance ($0-20m$, $20-40m$, ...), orientation index (parallel opposite, parallel same, ...), speed ($0-2.5m/s$, $2.5-5m/s$, ...) and action (turn left, turn right, accelerate, ...). GPT-4o performs text-to-text translation, converting traffic scene description into a YAML-like structured format through in-context learning~\cite{min2022rethinking}. Fig.~\ref{fig:generation_pipeline} provides a brief example of our prompt system.

Given $z$ and map $m$, the generative model produces a traffic scenario $\tau$ and consists of four modules:

- \textbf{Map Encoder}: processes a map region $m=(v_1, ..., v_S)$ into map features $F=(f_1, ..., f_S)$, using MCG blocks~\cite{varadarajan2022multipath++}. 

- \textbf{Agent Query Generator}: converts the structured representation $z_i$ of agent $i$ into agent query $q_i$ with a sinusoidal position encoding and a learnable query vector.

- \textbf{Generative Transformer}: models agent-map and agent-agent interactions with input map features $F=(f_1, ..., f_S)$ and agent queries $Q = (q_1, ..., q_N)$, where $N$ is the number of agents. These are fed into transformer layers for updated agent features $Q^{*}$.

- \textbf{Scene decoder}: predicts each agent's position, attributes and motion for each agent feature $q^{*}_i$ using MLP, following LCTGen~\cite{tan2023language}. For the i-th agent, it computes agent positions $\hat{p}_i \in \mathbb{R}^S$ as $\hat{p}_i = softmax(e^{agent}_i \times [e^{lane}_1, ..., e^{lane}_S]^T)$, where $e^{agent}_i$ and $e^{lane}_j$ are agent and lane map mask embeddings generated by Generative Transformer and Map Encoder, respectively. Agent attributes (heading, velocity, size, position shift) are modeled with a K-way Gaussian Mixture Model (GMM), where $[\mu_i, \Sigma_i, \pi_i] = MLP(q^i)$. Future motion over $T-1$ steps is predicted by generating $K'$ potential trajectories $\{pos^{2:T}_{i, k}, prob_{i, k}\}^{K'}_{k=1} = MLP(q^*_i)$. For each time step $t$, the position $pos^{t}_{i, k} = (x, y, \theta)$ includes the agent's position $(x, y)$ and heading $\theta$. The agent states $s_{1:N}$ are then composed of their position, attributes, and motion.

For training, we minimize $\mathcal{L}{\text{position}}$ (cross-entropy), $\mathcal{L}{\text{attr}}$ (GMM NLL), and $\mathcal{L}_{\text{motion}}$ (MSE), following LCTGen~\cite{tan2023language}.
During inference, we sample the most probable values from the predicted position, attribute, and motion distributions for each agent across $T$ timestamps, generating agent statuses $s^{i}_{1:N}$, which form the final scenario $\tau = (m, s_{1:N})$.

\begin{figure*}[t]
    \centering
    \includegraphics[width=0.85\textwidth]{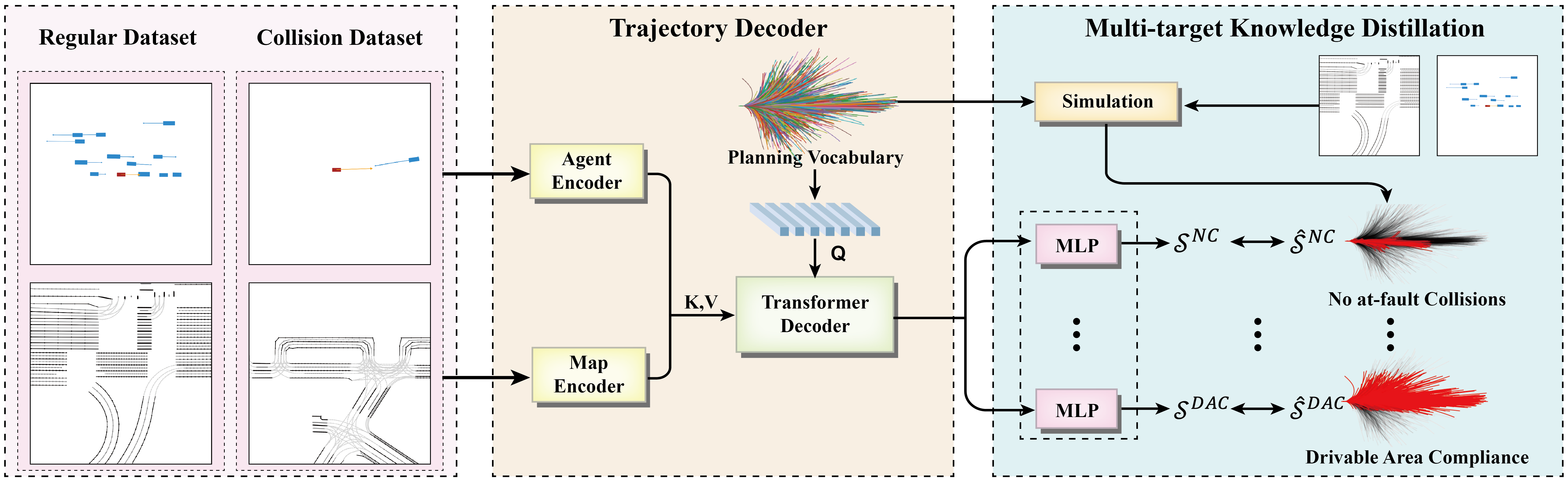} 
    \caption{SafeFusion aims to improve the safety of neural planners by integrating collision data into training, using a planning vocabulary and Multi-Target Hydra-Distillation to address the challenge of unavailable collision-avoidance trajectories in collision data.}
    \label{fig:planning}
    \vspace{-2em}
\end{figure*}

\subsubsection{\textbf{Filtering}}
\label{sec:filtering}
After the Generator's inference, the model produces a low proportion of collision scenarios where vehicles follow traffic rules. To ensure high-quality collision scenarios, we establish a set of rules that assess the generated vehicle trajectories within a given map region $m$. The entire filtering consists of two steps: \textbf{(1) Lane Compliance and Collision Filtering} and \textbf{(2) Collision Avoidance Feasibility Check}. These steps ensure that only meaningful collision scenarios are retained in our dataset.

\subsection*{Step 1: Lane Compliance and Collision Filtering}
Recall that the map region $m$ consists of lane segments $v_j$, each defined by its endpoints $(x_{1j}, y_{1j})$ and $(x_{2j}, y_{2j})$. This step filters vehicle trajectories based on three key criteria:

\textbf{Lane Adherence Compliance} ensures the vehicle stays within a lateral deviation limit $D_{thres}$ from the lane centerline. At each time step $i$, the minimum distance $d_i$ between the vehicle's position $(x_i, y_i)$ and the nearest lane segment $v_{j^*}$ is computed:
\begin{equation*}
d_i = \min_{v_j \in m} \{Distance((x_i, y_i), v_j) \}
\end{equation*}
where $Distance$ calculates the perpendicular distance from a point to a line segment. The vehicle satisfies this criterion if, for every time step $i$, the condition $d_i \leq D_{thres}$ holds.

\textbf{Lane Direction Alignment} ensures the vehicle's heading remains aligned with the lane centerline within an allowable angular deviation of $\Theta_{thres}$. At each time step $i$, the closest lane segment $v_{j^*}$ is identified, and the alignment angle $\theta_i$ between the vehicle's heading vector $v_{ego, i}$ and the lane direction vector $v_{lane, j^*}$ is computed as:
\begin{equation*}
\theta_i = \arccos \frac{v_{ego, i}\cdot v_{lane, j^*}}{||v_{ego, i}|| ||v_{lane, j^*}||}.
\end{equation*}
The vehicle satisfies this criterion if, for every time step $i$, the angular deviation $\theta_i \leq \Theta_{thres}$.

\textbf{Collision Involvement} detects potential collisions between the ego vehicle and other agents based on their respective trajectories, headings, and physical dimensions. This is implemented by checking if the bounding boxes of the ego vehicle and other agents intersect. If an intersection is detected, it is considered a collision.

\subsection*{Step 2: Collision Avoidance Feasibility Check}
\label{sec:filtering_step_2}
To ensure that the generated collision scenarios can potentially be avoided by the ego vehicle through trajectory planning, we utilize the PDM simulator~\cite{dauner2023parting} to evaluate feasible avoidance trajectories. The simulator takes an input trajectory to control the ego vehicle and transforms it into a dynamically feasible one by integrating it with the historical ego status. Following VADv2~\cite{chen2024vadv2} and HydraMDP~\cite{li2024hydra}, we build a fixed planning vocabulary to discretize the continuous action space. Specifically, we randomly sample 700,000 trajectories from the nuPlan database~\cite{caesar2021nuplan}, where each trajectory $T_i (i=1,..., k)$ consists of 40 timestamps, representing a 4-second horizon. K-means clustering is applied to obtain $k$ cluster centers, forming the vocabulary $\mathcal{V}_k$. This discretization effectively reduces the continuous action space into a manageable set of representative trajectories.

Using this vocabulary, the PDM simulator scores each trajectory in the generated scenarios. Scenarios where all trajectories lead to collisions are discarded, retaining scenarios that have at least one feasible collision-avoidance trajectory, as shown in Step 2 of the Filtering section in Fig.~\ref{fig:generation_pipeline}.

Only samples that simultaneously meet these two filtering steps are retained in our dataset, while all others are discarded. After these automated steps, we manually review the entire constructed collision dataset to further ensure its quality, resulting in the final \textit{Collision2k} dataset. This dataset includes scenarios where other vehicles collide with the ego vehicle from multiple directions, such as from the left front, directly in front, and right front of the ego vehicle, both at intersections and in open road areas.

\subsection{SafeFusion}
\label{sec:planning}


To address the absence of successful collision avoidance trajectories in the collision dataset and explicitly train the planner to avoid collisions, we use the trajectory vocabulary mentioned in Section~\ref{sec:filtering_step_2} and adopt Hydra-MDP\cite{li2024hydra}’s planning trajectory vocabulary decoder and Multi-target Hydra-Distillation strategy, while removing its imitation learning component. This modification overcomes Hydra-MDP’s limitation, where the lack of collision-avoidance trajectories in collision data hindered effective training. As shown in Fig.~\ref{fig:planning}, our architecture takes inputs from both the real-world regular dataset \( D_R \) and the collision dataset \( D_C \). We first extract environmental features \( F_{env} \) using encoders that process the ego vehicle's status, agent bounding boxes, attributes, and the map. The features are then used as the key and value for the Trajectory Transformer Decoder.


\textbf{Trajectory Decoder} generates possible future trajectories for the ego vehicle using the same vocabulary $\mathcal{V}_k$ as in Section~\ref{sec:filtering_step_2}. Each trajectory in $\mathcal{V}_k$ is embedded into a latent space via an MLP, creating a set of latent queries. These queries are processed through multiple layers of Transformer decoders, where they attend to the previously computed environmental features $F{env}$. The output is a refined set of trajectory predictions $\mathcal{V}^{'}_k$ that consider the ego vehicle's status, surrounding agents, and map constraints. 

\textbf{Multi-target Knowledge Distillation} can efficiently teach the planner to avoid obstacles using simulation scores as supervisory signals.
The distillation process expands the learning target in two steps: (1) 
\textbf{Generating Ground Truth Scores}: For each scenario in the training dataset, we run offline simulations using the PDM simulator~\cite{dauner2023parting} to generate ground truth scores $\{{\hat{\mathcal{S}}^{m}_{i} | i = 1,..., k}\}^{|M|}_{m=1}$ for each trajectory in the planning vocabulary. These scores are based on a variety of closed-loop metrics $M$, such as safety, efficiency, and comfort, as defined in the Navsim Challenge~\cite{Dauner2024ARXIV}. (2) \textbf{Prediction Heads}: During training, the latent vectors $\mathcal{V}^{'}_k$ generated by the Trajectory Decoder are passed through a set of Prediction Heads, which predict the simulation scores $\{{\mathcal{S}^{m}_{i} | i = 1,..., k}\}^{|M|}_{m=1}$ for each trajectory in the vocabulary. The model is trained using binary cross-entropy loss to align predicted scores with the ground truth:
\vspace{-1mm}
{
\begin{equation*}
\mathcal{L} = - \sum_{m, i} \hat{\mathcal{S}}^{m}_{i} \log{\mathcal{S}^{m}_{i}} + (1-\hat{\mathcal{S}}^{m}_{i})\log{(1-\mathcal{S}^{m}_{i})}.
\vspace{-5pt}
\end{equation*}
}

To enhance planner performance, our training process incorporates both real-world data $D_R$ and collision data $D_C$. In each iteration, batches are randomly sampled from $D_R$ or $D_C$ with probabilities $p_{R}$ and $p_{C}$, and the corresponding loss, $\mathcal{L}_R$ or $\mathcal{L}_C$, is weighted by $w_R$ or $w_C$. This ensures a balanced contribution from both datasets during training. The pseudo-code~\ref{algo} summarizes the process.
\vspace{-10pt}
\begin{algorithm}
\caption{Training with $D_R$ and $D_C$}
\begin{algorithmic}[1]
\While{training not completed}
    \State Generate random number $r \in [0, 1]$
    \If{$r < p_{R}$} 
        \State $batch \gets$ Randomly select a batch from $D_R$
        \State $\mathcal{L}_R \gets \text{loss\_function}(model, batch)$
        \State $\mathcal{L} \gets w_R \times \mathcal{L}_R$
    \Else
        \State $batch \gets$ Randomly select a batch from $D_C$
        \State $\mathcal{L}_C \gets \text{loss\_function}(model, batch)$
        \State $\mathcal{L} \gets w_C \times \mathcal{L}_C$
    \EndIf
    
    \State Compute gradients $\nabla \mathcal{L}$ and update parameters
\EndWhile
\end{algorithmic}
\label{algo}
\end{algorithm}

The architecture of SafeFusion and the design of its loss function naturally and seamlessly unify regular data and collision data for planner training, enhancing its safety.

\section{EXPERIMENT}
In this section, we first introduce our dataset and the evaluation metrics. Next, we present the experimental results for collision scenario generation and evaluate the performance of our planner trained with both regular and collision data.

\subsection{Dataset and Metrics}
\textbf{Dataset.} We use the OpenScene~\cite{openscene2023} dataset, a compact version of nuPlan~\cite{caesar2021nuplan}, as $D_R$, focusing on scenarios with changes in intention where the ego vehicle's historical data cannot be extrapolated. It is divided into 1192 training (Navtrain) and 136 testing (Navtest) samples. Our \textit{SafeFusion} framework effectively processes collision data, regardless of whether it is sourced from real-world environments or generated synthetically. Due to the lack of real-world collision datasets, we use the synthetic \textit{Collision2k} dataset as $D_C$, which contains 1,902 scenarios randomly split into 1,702 for training and 200 for testing.

\textbf{Metrics.} We evaluate scene realism using the Maximum Mean Discrepancy (MMD) score~\cite{gretton2012kernel} for agents’ positions and attributes, following~\cite{feng2023trafficgen, tan2023language}. MMD quantifies the distributional difference between generated and real data as \( \text{MMD}^2 = \mathbb{E}[k(x, x')] + \mathbb{E}[k(y, y')] - 2\mathbb{E}[k(x, y)] \), where \( k(\cdot, \cdot) \) is a kernel function (e.g., Gaussian). Lower MMD values indicate higher realism. Motion realism is assessed using mean average displacement error (mADE) and mean final displacement error (mFDE). We match real and generated agents using the Hungarian algorithm and compute relative trajectories based on initial positions and headings.

For the performance of planning, we use the PDM score in the Navsim Challenge~\cite{Dauner2024ARXIV}, which can be formulated as:
{
\begin{equation*}
PDM_{score} = NC \cdot DAC \cdot DDC \cdot \frac{(5 \cdot TTC + 2 \cdot C + 5 \cdot EP)}{12}
\end{equation*}
}
The sub-metrics NC, DAC, TTC, C, EP correspond to the No At-fault Collisions, Drivable Area Compliance, Time to Collision, Comfort, and Ego Progress. 

We choose non-reactive closed-loop evaluation because open-loop evaluation is insufficient for testing planning algorithms effectively~\cite{li2024ego, dauner2023parting}, and reactive closed-loop simulation cannot generate safety-critical scenarios~\cite{caesar2021nuplan}. Non-reactive closed-loop evaluation provides a controlled environment with fixed agent behavior, allowing more thorough performance assessment under challenging conditions.

\begin{table}[t]
\scriptsize
\centering
\begin{tabularx}{0.48\textwidth}{c|>{\centering\arraybackslash}X>{\centering\arraybackslash}X>{\centering\arraybackslash}X>{\centering\arraybackslash}X|>{\centering\arraybackslash}X>{\centering\arraybackslash}X}
\toprule
& \multicolumn{4}{c|}{Initialization (MMD) $\downarrow$} & \multicolumn{2}{c}{Motion $\downarrow$ } \\ \cline{2-7} 
& Position   & Heading  & Speed   & Size    & mADE    & mFDE      \\ \midrule
TrafficGen~\cite{feng2023trafficgen} & 0.2025 & 0.1927 & 0.2482 & 0.2164 & 1.911 & 6.494 \\
LCTGen~\cite{tan2023language} & 0.0621 & 0.1095 & \textbf{0.0722} & 0.1205 & 1.241 & 2.564 \\
\rowcolor{lgreen}
\textbf{CollisionGen (Ours)} & \textbf{0.0612} & \textbf{0.1037} & 0.0743 & \textbf{0.1196} & \textbf{1.189} & \textbf{2.454} \\ \bottomrule
\end{tabularx}
\vspace{-3pt}
\caption{Traffic scenario generation realism evaluation.}
\vspace{-10pt}
\label{tab:reconstruction}
\end{table}

\begin{table}[t]
\centering
\begin{tabular}{c|cc}
\toprule
                & \# After Step 1 $\uparrow$ & \# After Step 2 $\uparrow$ \\ \midrule
Crash Report~\cite{crashreport}   &            888                  &                  517                \\
Concise prompts &                  1837            &                  1672                \\
Our prompts     &                 2268             &                  1902               \\ \bottomrule
\end{tabular}
\caption{Successful collision scenarios after two filtering steps from 53,510 total scenarios. Step 1 is Lane Compliance and Collision Filtering. Step 2 is Collision Avoidance Feasibility Check. }
\label{tab:prompt}
\vspace{-7mm}
\end{table}

\textbf{Implementation Details.} Both the data generator and planner are trained and evaluated using 8 NVIDIA V100 GPUs. The data generator is configured with a maximum of $S = 384$ lanes and $N = 32$ vehicles per scene, simulating $T = 50$ timesteps at $10$ fps. The map dataset $M$ is built from the OpenScene training split, with attribute GMMs using $K = 5$ components and motion prediction using $K^{'} = 12$ modes. The generator is trained for $100$ epochs using AdamW~\cite{loshchilov2017decoupled}, Cosine Annealing~\cite{loshchilov2016sgdr}, a $3e-4$ learning rate, and a batch size of $512$. Filtering thresholds are set at $D_{thres} = 3m$ and $\Theta_{thres} = 10^\circ$.
In the planning experiments, we first train our modified Hydra-GT model~\cite{li2024hydra} following the original paper's settings, then fine-tune it with both the OpenScene dataset and the \textit{Collision2k} dataset using AdamW and Cosine Annealing for 100 epochs, with an initial learning rate of 2e-4 and a batch size of 256.

\vspace{-5pt}
\subsection{Evaluation on Collision Scenario Generation}
\subsubsection{Baselines} For the traffic generation experiment, we compare against two SOTA traffic generation methods, TrafficGen~\cite{feng2023trafficgen} and LCTGen~\cite{tan2023language} under same settings as ours.

\subsubsection{Quantitative Results}
Table~\ref{tab:reconstruction} compares the realism of generated traffic scenarios across different methods. Our approach achieves the lowest MMD values for initial position, heading, and size, while maintaining competitive speed initialization. Additionally, it outperforms both baselines in motion quality, achieving the lowest mADE and mFDE.

\begin{table*}[t]
\begin{tabularx}{\textwidth}{c|c|cc|>{\centering\arraybackslash}X>{\centering\arraybackslash}X>{\centering\arraybackslash}X>{\centering\arraybackslash}X>{\centering\arraybackslash}X>{\centering\arraybackslash}X>{\centering\arraybackslash}X}
\toprule
\multicolumn{2}{c|}{Collision2k}  & Imi. & KD & \textbf{NC} & DAC & DDC & EP & \textbf{TTC} & COMF & Total \\ 
\midrule
\multirow{2}{*}{Rule-based} & IDM~\cite{treiber2000congested}   & & & 0.390 & 0.870 & 0.952 & 0.326 & 0.185 & 0.840 & 0.250 \\
                                     & PDM-closed~\cite{dauner2023parting} & & & 0.400 & 0.925 & 1.0 & 0.368 & 0.195 & 0.850 & 0.283  \\
\midrule
\multirow{5}{*}{Learning-based} & PlanTF~\cite{cheng2024rethinking}  & \checkmark &   & 0.280 & 0.710 & 0.942 & 0.182 & 0.135 & 0.995 & 0.147   \\
                                        & Hydra-GT~\cite{li2024hydra}    & \checkmark & \checkmark   & 0.480 & 0.775 & 0.968 & 0.355 & 0.315 & 0.375 & 0.266  \\
                                        & Hydra-GT~\cite{li2024hydra}     & & \checkmark  & 0.530 & 0.845 & 0.975 & 0.406 & 0.440 & 0.140 & 0.336  \\
                                        \rowcolor{lgreen} & SafeFusion (Ours)   &  & \checkmark & 0.625 & 0.895 & 0.995 & 0.491 & 0.515 & 0.175 & \textbf{0.415}  \\
\bottomrule
\end{tabularx}
\caption{Evaluation on collision Scenarios from the Collision2k test set. ``Imi.'' stands for imitation learning, and ``KD'' stands for knowledge distillation. These abbreviations also apply to the table below. Our method significantly enhances the planner's performance in collision corner cases and surpasses previous approaches by a notable margin in PDM scores. }
\label{tab:collision_test}
\vspace{-3em}
\end{table*}

\begin{table}[t]
\tiny
\centering
\begin{tabularx}{0.485\textwidth}{c|cc|>{\centering\arraybackslash}X>
{\centering\arraybackslash}X>{\centering\arraybackslash}X>{\centering\arraybackslash}X>{\centering\arraybackslash}X>{\centering\arraybackslash}X>{\centering\arraybackslash}X}
\toprule
OpenScene  & Imi. & KD & NC & DAC & DDC & EP & TTC & C & Total \\  \midrule
Hydra-GT~\cite{li2024hydra}    & \checkmark & \checkmark   & 0.981 & 0.914 & 1.0 & 0.761 & 0.948 & 1.0 & 0.831 \\
Hydra-GT~\cite{li2024hydra}     & & \checkmark  & 0.982 & 0.924 & 1.0 & 0.754 & 0.951 & 1.0 & 0.833 \\
\rowcolor{lgreen} SafeFusion (Ours)   &  & \checkmark & 0.980 & 0.925 & 1.0 & 0.755 & 0.947 & 1.0 & \textbf{0.832} \\ 
\bottomrule
\end{tabularx}
\caption{Evaluation on regular driving Scenarios from the OpenScene test set. Our method preserves the planner's performance in regular driving scenarios, even after targeted training focused on collision scenarios.}
\label{tab:openscene}
\vspace{-3em}
\end{table}

\begin{table}[t]
\scriptsize
\centering
\begin{tabularx}{0.485\textwidth}{c|>{\centering\arraybackslash}X>{\centering\arraybackslash}X>{\centering\arraybackslash}X>{\centering\arraybackslash}X>{\centering\arraybackslash}X>{\centering\arraybackslash}X>{\centering\arraybackslash}X}
\toprule
Hard samples  & NC & DAC & DDC & EP & TTC & C & Total \\  \midrule
Hydra-GT~\cite{li2024hydra}      & 0.631 & 0.824 & 1.0 & 0.440 & 0.007 & 0.99 & 0.264 \\
\rowcolor{lgreen} SafeFusion (Ours)   & 0.690 & 0.836 & 1.0 & 0.474 & 0.308 & 1.0 & \textbf{0.400} \\ \bottomrule
\end{tabularx}
\caption{Performance on OpenScene hard samples.}
\label{tab:openscene_failure}
\vspace{-20pt}
\end{table}

\begin{table}[t]
\scriptsize
\centering
\begin{tabularx}{0.485\textwidth}{c|c|>{\centering\arraybackslash}X>{\centering\arraybackslash}X>{\centering\arraybackslash}X>{\centering\arraybackslash}X>{\centering\arraybackslash}X>{\centering\arraybackslash}X>{\centering\arraybackslash}X}
\toprule
Collision2k  & ratio & NC & DAC & DDC & EP & TTC & C & Total \\  \midrule
Hydra-GT~\cite{li2024hydra} & - & 0.480 & 0.775 & 0.968 & 0.355 & 0.315 & \textbf{0.375} & 0.266 \\ \midrule
Concatenate& - & 0.520 & 0.885 & 0.985 & 0.403 & 0.425 & 0.295 & 0.346 \\ \midrule
Acc. Grad. & 10:1 & 0.600 & 0.880 & 0.995 & 0.448 & 0.500 & 0.135 & 0.380 \\ \midrule
\multirow{3}{*}{Random} & 5:1 & \textbf{0.655} & 0.880 & 0.995 & 0.478 & \textbf{0.555} & 0.170 & 0.411 \\
  & 15:1 & 0.605 & \textbf{0.915} & 0.995 & 0.469 & 0.520 & 0.135 & 0.404 \\
 & 10:1 & 0.625 & 0.895 & 0.995 & \textbf{0.491} & 0.515 & 0.175 & \textbf{0.415} \\ \bottomrule
\end{tabularx}
\caption{Ablation study on Collision2k test set.}
\label{tab:collision_ablation}
\vspace{-18pt}
\end{table}

\begin{table}[t]
\scriptsize
\centering
\begin{tabularx}{0.485\textwidth}{c|c|>{\centering\arraybackslash}X>{\centering\arraybackslash}X>{\centering\arraybackslash}X>{\centering\arraybackslash}X>{\centering\arraybackslash}X>{\centering\arraybackslash}X>{\centering\arraybackslash}X}
\toprule
Openscene  & ratio & NC & DAC & DDC & EP & TTC & C & Total \\  \midrule
Hydra-GT~\cite{li2024hydra} & - & \textbf{0.981} & 0.914 & 1.0 & \textbf{0.761} & 0.948 & 1.0 & 0.831 \\ \midrule
Concatenate & - & 0.978 & 0.919 & 1.0 & 0.759 & 0.945 & 1.0 & 0.830 \\ \midrule
Acc. Grad. & 10:1 & 0.979 & 0.925 & 1.0 & 0.738 & 0.949 & 1.0 & 0.825 \\ \midrule
\multirow{3}{*}{Random} & 5:1 & 0.967 & 0.920 & 1.0 & 0.711 & 0.929 & 1.0 & 0.802 \\
  & 15:1 & 0.980 & \textbf{0.926} & 1.0 & 0.756 & \textbf{0.949} & 1.0 & \textbf{0.835} \\
 & 10:1 & \textbf{0.980} & 0.925 & 1.0 & 0.755 & 0.947 & 1.0 & 0.832 \\ \bottomrule
\end{tabularx}
\caption{Ablation study on OpenScene test set.}
\label{tab:openscene_ablation}
\vspace{-30pt}
\end{table}

\subsubsection{Ablation study with different prompts} Table~\ref{tab:prompt} presents the number of collision scenarios generated by different prompt sets when applied across the entire dataset. As shown, our tailored prompts produce more crash scenarios compared to both the prompts used in LCTGen’s crash reports~\cite{crashreport} and a simplified version of our own prompts. The simplified version is derived using ChatGPT to streamline our original prompts. The generated scenarios are then utilized for training and evaluating planning models.

\vspace{-3pt}
\subsection{Evaluation on Planners}
\subsubsection{Baselines}
We compare our method, SafeFusion, with several baselines:
i) \textbf{IDM}~\cite{lopez2018microscopic} is a mathematical model for simulating driving behavior, based on parameters like acceleration and desired speed.
ii) \textbf{PDM}\cite{dauner2023parting}, a rule-based winner of the 2023 nuPlan Planning Challenge. PDM-closed ensembles IDM\cite{lopez2018microscopic} with different hyperparameters to optimize performance.
iii) \textbf{PlanTF}~\cite{cheng2024rethinking} is a transformer based neural planner based purely on imitation learning.
iv) \textbf{Hydra-GT} is a modified version of Hydra-MDP~\cite{li2024hydra}, which uses ground-truth post-perception results instead of raw sensor data as input, while retaining the multi-target knowledge distillation architecture. We implement two versions: one combining imitation learning with knowledge distillation, and the other relying solely on knowledge distillation.

\subsubsection{Quantitative Results}

Table~\ref{tab:collision_test} shows the performance of different planners on \textit{Collision2k} test set. We see that: i) Rule-based methods score low as the malicious collision scenarios fall outside the scope of their predefined rules. ii) PlanTF and Hydra-GT still face unavoidable collisions due to a lack of similar training data. iii) Our method, incorporating collision data and employing a mixed training strategy with Knowledge distillation, significantly improves the planner's ability to handle corner cases. iv) Compared to PDM-closed, our method improves NC by \textbf{56.3}\%, TTC by \textbf{164.1}\%, and the overall score by \textbf{46.6}\%. It also shows a  \textbf{30.2}\% improvement in NOC, \textbf{63.5}\% in TTC, and \textbf{56.0}\% in the overall score compared to our backbone model Hydra-GT. 

Table~\ref{tab:openscene} shows the performance of different methods on the OpenScene test set. We observe that, after fine-tuning, our method is able to maintain the original performance of the backbone method on the OpenScene dataset. 


Our model's performance is also tested on Hydra-GT’s failure cases, particularly those with $NC=0$ or $TTC=0$. Table~\ref{tab:openscene_failure} shows the evaluation on these difficult samples from the OpenScene dataset. After fine-tuning, our model demonstrated a \textbf{30}\% improvement in collision avoidance.

\begin{figure}[ht!]
    \centering
    \includegraphics[width=0.44\textwidth]{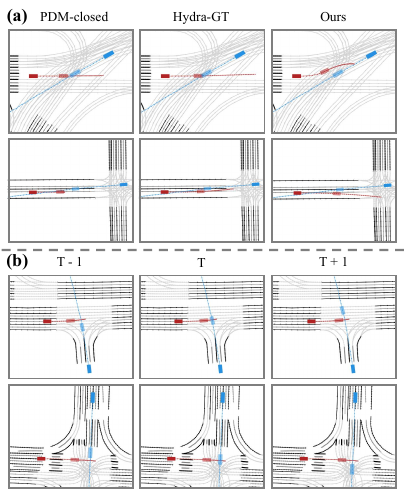} 
    \caption{Visualizations of planners.}
    \label{fig:comparison}
    \vspace{-20pt}
\end{figure}

\subsubsection{Qualitative Results}
Fig.~\ref{fig:comparison} (a) highlights some scenarios that our method outperforms the baselines. Fig.~\ref{fig:comparison} (b) shows how our planner progresses over time. After fine-tuning, the planner: i) slows down to yield to other vehicles and avoid crash, ii) is capable of collision avoidance through lane changing, iii) can consider the future trajectories of other vehicles to select collision-free routes, although at times the chosen route does not align with the lanes.

\subsubsection{Ablation Study}
Table~\ref{tab:collision_ablation} and Table~\ref{tab:openscene_ablation} show the ablation study on training strategies and data ratios for batch selection. ``Concatenate'' directly merges two datasets. ``Acc. Grad.'' with ratio 10:1 accumulates gradients of 10 batches from OpenScene and 1 batch from Collision2k before back propagation. In ``Random'', batches are randomly selected from the two datasets based on a specific ratio. We observe that: i) Random batch selection outperforms both concatenation and gradient accumulation. ii) The ratio in random batch selection is crucial—too small shifts the model toward the collision domain, degrading original dataset performance, while too large diminishes the impact of collision data, limiting the algorithm's ability to handle dangerous cases.




\section{CONCLUSION}
In this work, we tackle the challenges of improving autonomous vehicle safety in unpredictable and hazardous scenarios, focusing on the scarcity of collision data and its integration into planner training. We introduce CollisionGen, a scalable pipeline for generating realistic and diverse collision scenarios using natural language prompts and a generative model, resulting in the Collision2k dataset. Additionally, we propose SafeFusion, a training framework that addresses challenges such as over-reliance on imitation learning, data imbalance, and domain discrepancies, ultimately enhancing collision avoidance learning and overall planner performance. This approach paves the way for more reliable and adaptable autonomous systems capable of navigating complex environments safely and efficiently. A limitation of this work is that it focuses primarily on collision scenarios, with expansion to other hazardous situations as a direction for future work.

\bibliographystyle{IEEEtran}
\bibliography{IEEEabrv,main.bbl}

\end{document}